\documentclass[letterpaper, 10 pt, conference]{ieeeconf} 

\IEEEoverridecommandlockouts                              

\newcommand{\RNum}[1]{\uppercase\expandafter{\romannumeral #1\relax}}

\usepackage{graphics} 
\usepackage{graphicx}
\graphicspath{{./figure/}}
\usepackage{subcaption}
\usepackage[font=small]{caption}

\usepackage[noadjust]{cite}

\usepackage{hyperref}
\usepackage[ruled,linesnumbered,vlined]{algorithm2e}
\usepackage{amsmath} 
\usepackage{amssymb}  
\usepackage{bm}
\usepackage{romannum}
\usepackage{booktabs}
\usepackage{color}
\usepackage{xcolor}
\usepackage{enumerate}
\usepackage{svg}
\usepackage{scalerel}
\newcommand\Tau{\scalerel*{\tau}{T}}

\title{\LARGE \bf
Viko 2.0: A Hierarchical Gecko-inspired Adhesive Gripper \\with Visuotactile Sensor
}

\author{Chohei Pang$^{1}$, \textit{Student Member, IEEE}, Qicheng Wang$^{2}$, Kinwing Mak$^{1}$, \\Hongyu Yu$^{1}$, and Michael Yu Wang$^{1}$, \textit{Fellow, IEEE}
	\thanks{Research is supported by the Hong Kong Innovation and Technology
Fund (ITF) ITS-104-19FP. \textit{(Chohei Pang and Qicheng Wang contributed equally to this work.)} }
	\thanks{$^{1}$C. Pang, K. Mak, H. Yu (corresponding author), and M. Y. Wang (corresponding author), are with the Department of Mechanical and Aerospace Engineering, Hong Kong University of Science and Technology, Hong Kong (e-mail: chpangad; kwmakaf; hongyuyu; mywang@ust.hk).}
    \thanks{$^{2}$ Q. Wang is with Robotics and Autonomous Systems Thrust, Systems Hub, Hong Kong University of Science and Technology, Hong Kong (email: qwangco@ust.hk)}
    }

\SetKwInput{KwData}{Input}
\SetKwInput{KwResult}{Output}

\begin{document}

\maketitle
\thispagestyle{empty}
\pagestyle{empty}

\begin{abstract}
Robotic grippers with visuotactile sensors have access to rich tactile information for grasping tasks but encounter difficulty in partially encompassing large objects with sufficient grip force. While hierarchical gecko-inspired adhesives are a potential technique for bridging performance gaps, they require a large contact area for efficient usage. In this work, we present a new version of an adaptive gecko gripper called \textit{Viko} 2.0 that effectively combines the advantage of adhesives and visuotactile sensors. Compared with a non-hierarchical structure, a hierarchical structure with a multi-material design achieves approximately a 1.5 times increase in normal adhesion and double in contact area. The integrated visuotactile sensor captures a deformation image of the hierarchical structure and provides a real-time measurement of contact area, shear force, and incipient slip detection at 24 Hz. The gripper is implemented on a robotic arm to demonstrate an adaptive grasping pose based on contact area, and grasps objects with a wide range of geometries and textures.
\end{abstract}

\section{Introduction}
 
In nature, hierarchical structures have functional properties such as hydrophobicity and adhesion to surfaces \cite{bae201425th,koch2009hierarchically}. These structures can alter a material's properties and allow surface adaptiveness across multiple length scales \cite{gao2005mechanics,kim2007effect,bhushan2006adhesion}. In particular, in the gecko's toe, the hierarchical fibrillar structure can conform intimately to the surface at multiple levels, thus ensuring the presence of a strong van der Waals force \cite{Autumn2000}. The contact load between the toe and the surface is distributed evenly across millions of separate contacts \cite{russell2007insights,Autumn2006}. Although synthetic dry adhesives inspired by the hierarchical structure of the gecko's toe adopt its adaptability, they must also ensure a sufficient amount of real contact area to achieve high adhesion. Therefore, the sensing of contact information by hierarchical structures is essential for such structures to be implemented in the robotics field.

In \cite{Pang2021}, we presented \textit{Viko}, an adjustable gecko gripper with a visuotactile sensor. \textit{Viko} estimates the pixel-level contact area and shear force with high precision. However, like the conventional visuotactile sensor, it uses a piece of silicone rubber as a base material. It cannot adequately conform to the shape of the contact object without significant distortion. In contrast, a hierarchical structure can easily conform to the surface and yield a larger contact area.
\begin{figure}
	\centering
	\includegraphics[width=0.48\textwidth]{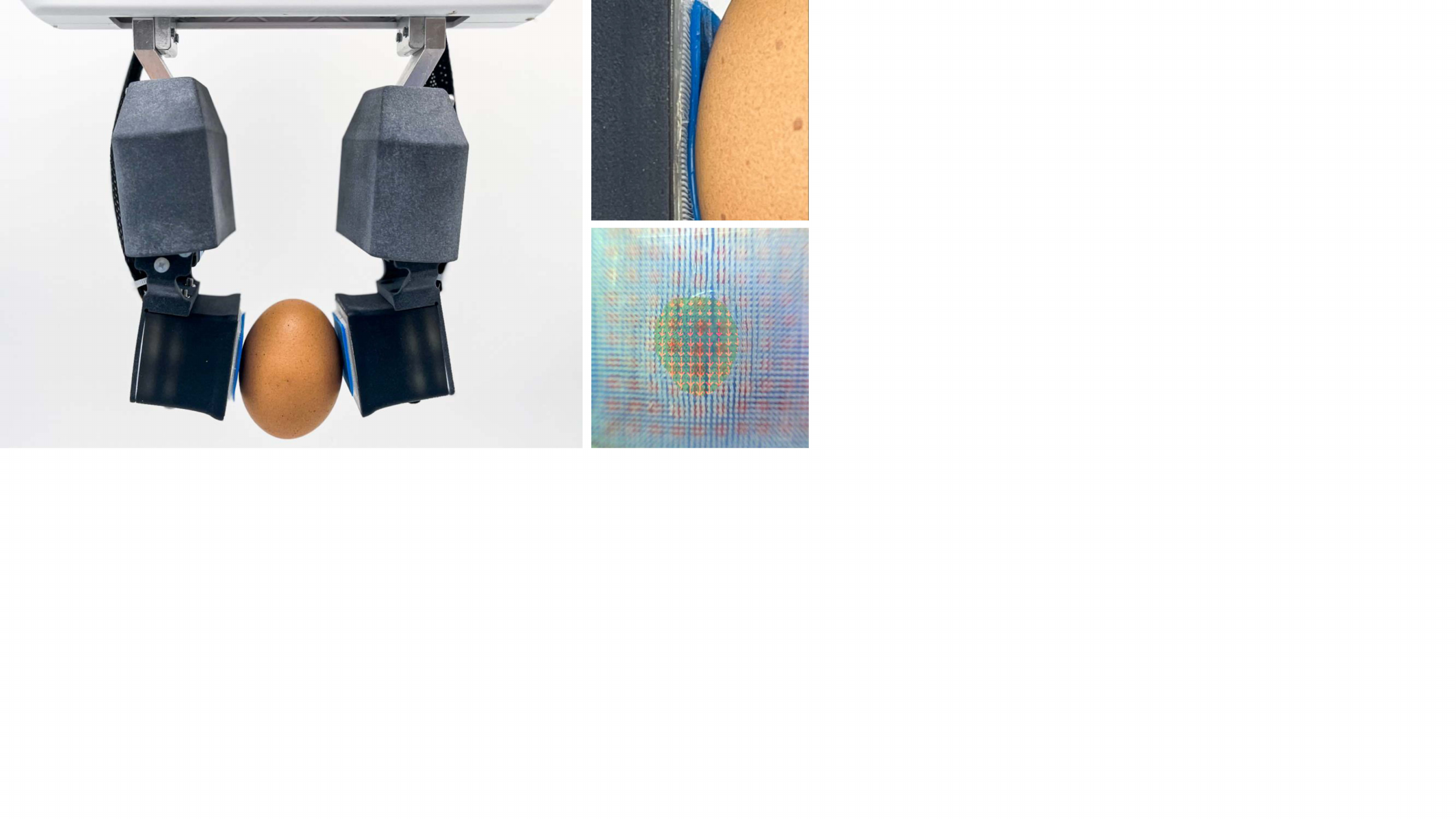}
	\caption{Left: The \textit{Viko} 2.0 adaptive gripper installed on a robotic arm and grasping an egg. Top-right: A close-up view of the hierarchical gecko-inspired adhesive adapting to the egg. Bottom-right: Real-time tracking of the hierarchical adhesive's contact area (green) and shear force (red arrow). }
	\vspace{-0.4cm}
	\label{fig: intro_graph}
\end{figure}
Herein, we present \textit{Viko} 2.0, which is designed to simultaneously adopt a high-performance hierarchical gecko-inspired adhesive and sense in real time the contact information, including contact area, shear force, and incipient slip, in a compact gripper design. The gripper fingers are mounted on a parallel gripper with controllable joints providing an extra degree of freedom. An image segmentation algorithm is used to determine the high-resolution contact region, while a feature tracking method can estimate the shear force and incipient slip. In addition, the multi-material design of the hierarchical structure ensures superior adhesive forces and contact area performance. 
In this paper, we present three critical contributions to the tactile sensing of hierarchical gecko-inspired adhesives, which enable gentle grasping of everyday objects: 
\begin{enumerate}
    \item A high-performance hierarchical gecko-inspired adhesive through the multi-material design of the suspension layer.
    \item Techniques for adding internal patterned markers to the hierarchical structure, thus enabling feature tracking for shear-induced membrane deformation.
    \item Sensing contact area, shear force, and incipient slip of the hierarchical structure.
\end{enumerate}
The paper is structured as follows: In Section \ref{sec: 2rela}, we review related works in the design of hierarchical gecko-inspired adhesive and tactile sensing technologies. We discuss the design, material selection, and fabrication of the adhesive structure and visuotactile sensor in Section \ref{sec: 3design} and describe the pipeline for sensing contact information in Section \ref{sec: sensing}. We describe experiments to evaluate the performance of the gripper in grasping daily objects in Section \ref{sec: exp}. Finally, we briefly discuss the contributions of the paper and future work in Section \ref{sec: conc}.

\section{Related Works}
\label{sec: 2rela}
\subsection{Hierarchical Gecko-inspired Adhesive}
In gecko hierarchical structures, the lamellae are split into hundreds of tiny setae, which can conform to the shapes of irregular-surface objects and thereby overcome the variation in surface alignment \cite{Autumn2002,Autumn2006,Autumn2006-2}. These structures achieve a large contact area and superior adhesion. Synthetic pillars with hierarchical structures also exhibit better adhesive performance than those with single-layer structures \cite{Wang2014,greiner2009experimental}.
Many synthetic hierarchical gecko-inspired structures are fibrillar, typically using a combination of a series of nanopatterning techniques with various micromanufacturing techniques \cite{northen2008gecko,jeong2009nanohairs,bhushan2012fabrication,murphy2009enhanced}. Asbeck et al. developed a hierarchical system consisting of a suspension of directional pillars and a top layer of wedge-shaped adhesives \cite{asbeck2009climbing}. Two layers were cast by the molds separately and bonded together after being fully cured. The adhesives were integrated into a climbing robot, which could climb a rough surface safely. 
\subsection{Gecko-inspired Adhesive Sensing}
Two critical parameters can be used to evaluate the performance of gecko-inspired adhesives: the contact area and the shear force. However, although numerous sensing methods, such as capacitance \cite{Wu2015,huh2018active}, optical \cite{Eason2015,Hawkes2015}, and resistance \cite{drotlef2017bioinspired} sensing, can monitor the shear force and contact area, few technologies can provide high-resolution measurements of these parameters using a monitor with a compact size. 
Previous studies have measured the stress distribution and contact area of a gecko toe using high-resolution sensors based on the optical principle of frustrated total internal reflection (FTIR) with a micro-textured pattern membrane. The contact information can be extracted from the area and the brightness when the membrane is in contact with the surface \cite{Eason2015}. In other studies, capacitance sensors have been integrated with a gecko-inspired adhesive on rigid tiles \cite{Wu2015} or thin-film backing \cite{huh2018active}. These sensors are arranged into a small number of sensing units to cover the entire adhesive area, and they can measure high-resolution shear and normal force. However, they can only determine whether the sensing unit is in contact with the surface or not. Therefore, to obtain the accurate contact area, a densely packed capacitance sensor matrix is required, which is costly and time-consuming to fabricate.
\subsection{Vision-based Tactile Sensor}
Recently, vision-based tactile sensors have been favored in robotics because of their ability to obtain high-resolution contact information while retaining a compact sensor size. These sensors capture the surface deformation of an elastomer in the form of an image and process this image with various model-based or learning-based algorithms.
Gelsight \cite{yuan2017gelsight}, Gelslim \cite{donlon2018gelslim}, and Digit \cite{lambeta2020digit} illuminate a reflective membrane surface using color LEDs to create colored, shaped shadow when an object is pressed against the membrane. The geometry and the contact information can be determined from the high-dimensional image data. Other sensors, such as Tactip \cite{ward2018tactip}, interpret the surface deformation by subsurface pin movement. From the deformation image, Tactip deduces the contact contour and forces. The Fingervision \cite{Du2021} sensor uses optical flow to extract the deformation vector field from a random dense pattern. Using a learning-based method, it achieves online contact feature extraction with high accuracy and frequency.

\section{Design and Fabrication}
\label{sec: 3design}
\subsection{Hierarchical Gecko-inspired Adhesive}
\subsubsection{\textbf{Hierarchical multi-material structure}}
Hierarchical structures can enhance adhesive performance. Although the gecko's hierarchical system contains only one substance, each upper fibril consists of hundreds of smaller branches, resulting in considerably softer lower branches. Therefore, the hardness differs between the top and bottom layers. We used materials of different hardness for the upper and lower layers to mimic the hardness variance between layers. For the microwedge structure layer, it has been reported that Mold Star 30 achieves a balance between surface energy and compliance, enabling high-performance gecko-inspired adhesion \cite{parness2009microfabricated}. Thus, Mold Star 30 with a Shore hardness of 30A was used for the microwedge structure layer. The material chosen for the pillar array layer has a wide Shore hardness range from 16A to 40A. This material was chosen based on normal adhesion, shear adhesion, and contact area tests.
\begin{figure*}[ht]
	\centering
	\vspace{0.1cm}
	\includegraphics[width=0.88\textwidth]{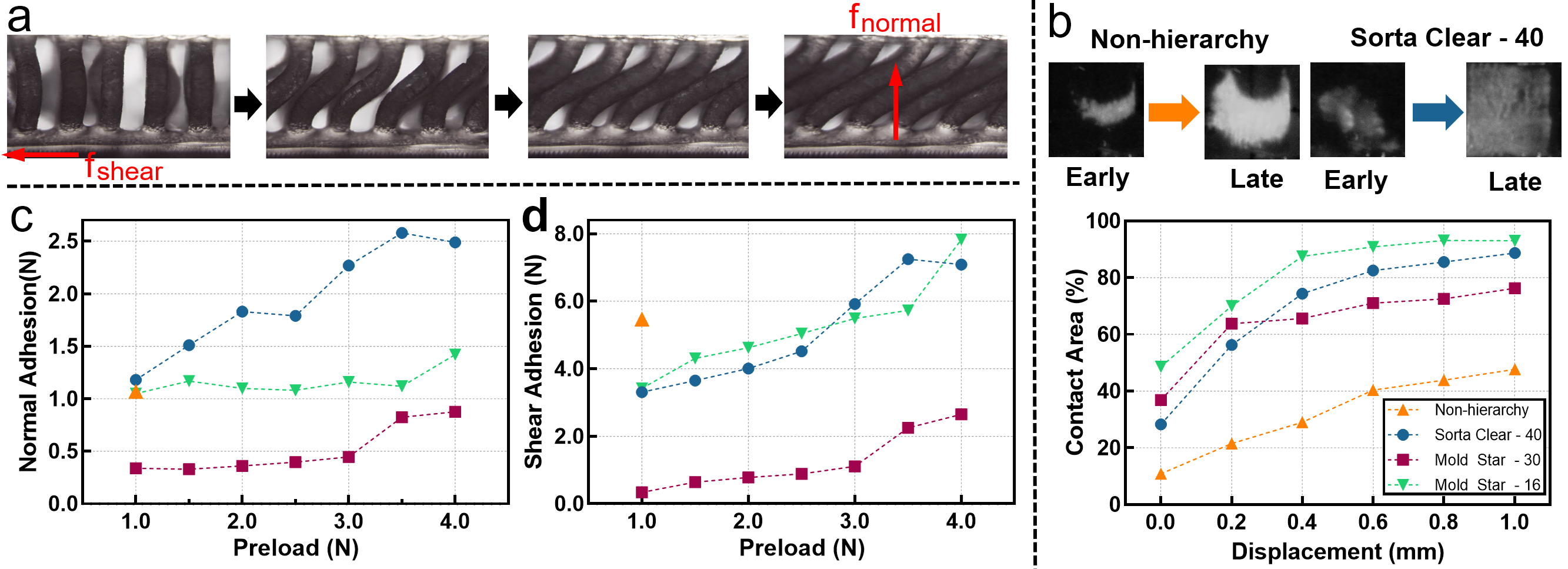}
	\caption{Performance evaluation and analysis of hierarchical structures with various hardness. (a) The hierarchical adhesive's pillar structure is subjected to shear loading, demonstrating the stages of conforming to the contact surface. (b) Contact area, as measured by FTIR, plotted against displacement in shear direction for three hierarchical structures of various hardness and a non-hierarchical sample. Top: The FTIR images of the contact area at early and late stages of the shear loading test. Normal adhesion $\mathrm{f_{normal}}$(c) and shear adhesion $\mathrm{f_{shear}}$ (d) versus preload plots of the same structures, showing that hierarchical structures achieve higher adhesive performances.}
	\label{fig: contactarea}
\end{figure*}
\subsubsection{\textbf{Performance test and material selection}}
The contact area was determined during displacement under 3 N loading and tested by a custom-built FTIR measurement apparatus \cite{Pang2021} that can measure the contact area at the micrometer scale \cite{han2005low}. When displacement is applied in the shear direction under loading, the pillar will first buckle, and bending along the shear direction then causes compliance to the contact surface, as shown in Fig. \ref{fig: contactarea}(a). Figure \ref{fig: contactarea}(b) shows the contact area measurement, which indicates that a hierarchical structure can enhance the contact area and lead to higher adhesive performance. The hierarchical structures with Mold Star 16 and Sorta Clear 40 achieved a higher contact area than that with Mold Star 30. We also conducted the standard load--drag--pull (LDP) tests as in \cite{wang2017adhesion} to evaluate the normal and shear adhesion. While Mold Star 16 and Sorta Clear 40 exhibited comparable shear adhesion, Sorta Clear 40 outperformed Mold Star 16 in normal adhesion.

The results reveal that a hierarchical structure with varying hardness outperforms that with a single hardness. Among the tested materials, we chose Sorta Clear 40 as it exhibited superior normal adhesive performance.
\subsubsection{\textbf{Dot Patterning}}
The sensing pattern in the conventional visuotactile sensor was attached to the fully cured elastomer membrane and then protected with another layer. It featured a complete surface connection between layers, ensuring a robust bond. However, the pillar-based hierarchical structure had a limited connection surface. The pillar needed to be inserted into the microwedge structure to achieve a stronger connection. We propose a method of adding sensing patterns to the inner layer of hierarchical structures in the uncured state. The colored dot-shaped pattern was stamped onto the back of the microwedge layer by an array of pins during the half-cured state. The viscosity of the elastomer is relatively low in this state, enabling the pattern to retain its position and shape for a more stable result. Moreover, the half-cured state enables the pillar structure to be immersed in the uncured microwedge structure and cured jointly.
\subsubsection{\textbf{Hierarchical Structure Fabrication Process}}
\begin{figure}
	\centering
	\includegraphics[width=0.48\textwidth]{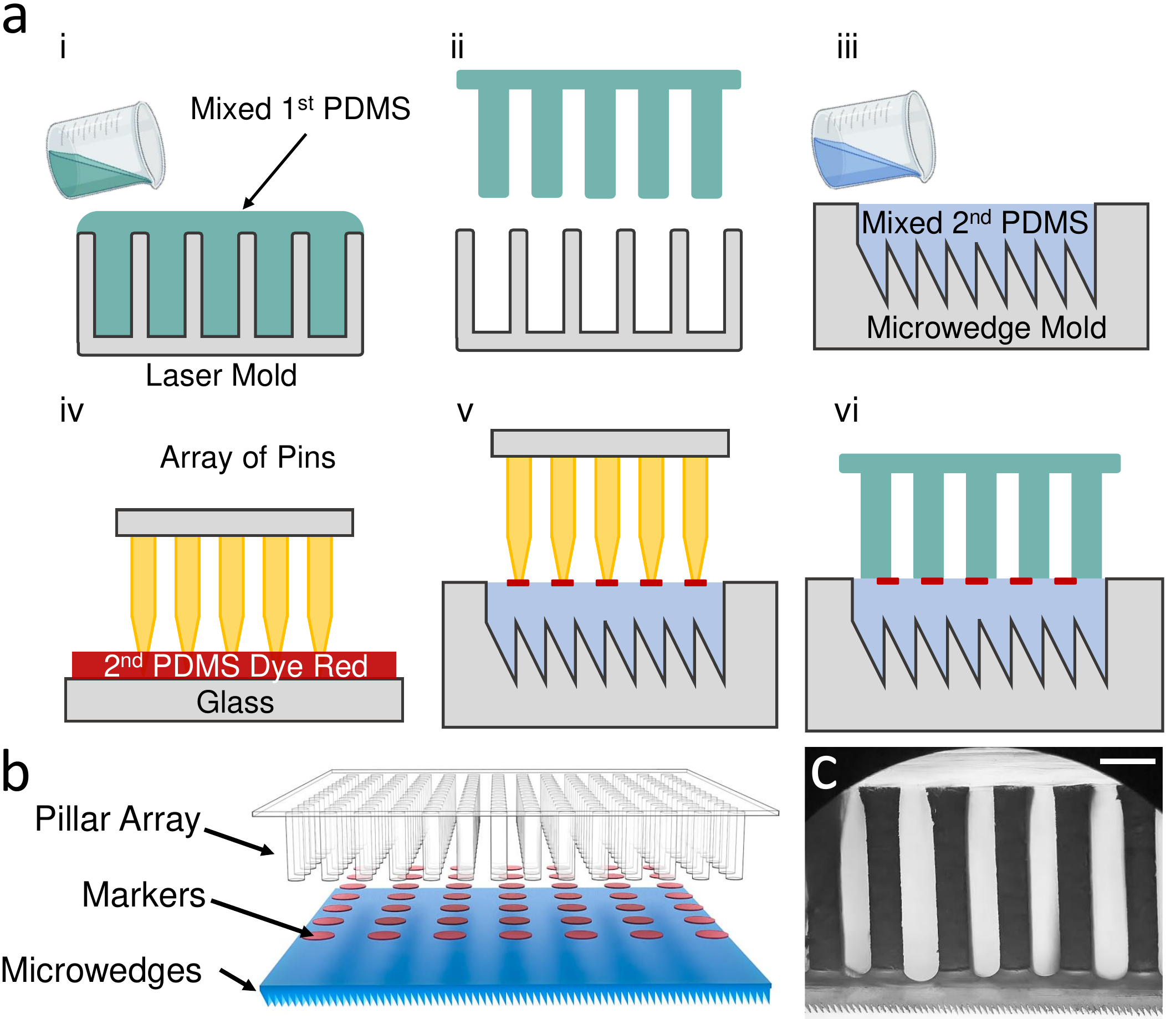}
	\vspace{-0.2cm}
	\caption{Hierarchical structure design and fabrication. (a) The manufacturing process of the hierarchical structure is illustrated in six simplified steps. (b) Computer-aided design (CAD)-rendered graph of the hierarchical structure in an exploded view. (c) Cross-section microscopic photo of the hierarchical structure. The white bar represents 500 $\mu m$.}
	\vspace{-0.4cm}
	\label{fig: fabrication}
\end{figure}
The fabrication process of the hierarchical structure is illustrated in Fig. \ref{fig: fabrication}(a). The structure can be separated into three layers, as illustrated in Fig. \ref{fig: fabrication}(b): pillar array, markers, and the microwedge structure. These three layers are tightly bonded together during the manufacturing processes, and a cross-section microscopic photo of the hierarchical structure is shown in Fig. \ref{fig: fabrication}(c). 

The laser mold had holes of 300 $\mu m$ diameter and 2 $mm$ height with a center-to-center distance of 600 $\mu m$. The first aliquot of PDMS was mixed and poured into the laser mold  and then vacuum-degassed for 2 minutes to remove bubbles and cured at 60$^\circ$C according to the material specifications. Once fully cured, it can be demolded and is ready for later processing. The microwedge mold had a wedge height of 100 $\mu m$ with a tip angle of 25.5$^\circ$ and a size of 3 $\times$ 3 $cm^2$. The second aliquot of PDMS was mixed and poured into the microwedge mold, and the remaining material was dyed red by the pigment (Silc Pig, Smooth-on) and poured on the glass substrate. Then, a 10 $\times$ 10 array of pins with spacing of 2.5 $mm$ was used to transfer the red dot markers onto the back of the microwedge. To control the size of the red markers by lowering the diffusion speed of the material, the second PDMS was half-cured at 60$^\circ$C for 6 minutes. Then, the pins were inserted into the microwedge mold and extracted, thus transferring the red markers onto the back surface of the microwedge. The pillar structure was also inserted into the microwedge mold, and the entire structure was fully cured at 60$^\circ$C for 2 hours. 
\subsection{Adaptive Gecko Gripper}
The adaptive gecko gripper functions similarly to the previous generation, with two opposing fingertips attached to a parallel gripper (Franka Emika), as seen in Fig. \ref{fig: design}(a). By connecting the parallel gripper and sensor module via a servo motor, the sensor module can be better aligned with the object based on contact feedback. The entire design becomes more compact and rigid yet provides robust performance. Two screws pass through the sensor module's inner and outer parts, as well as the connection arm, securing all components and enabling fast assembly. The protective shield protects the servo motor and connection arm and organizes the electric cables, resulting in a neat arrangement for easy maintenance.
\subsection{Sensor Module Design}
\begin{figure}
	\centering
	\vspace{0.1cm}
	\includegraphics[width=0.35\textwidth]{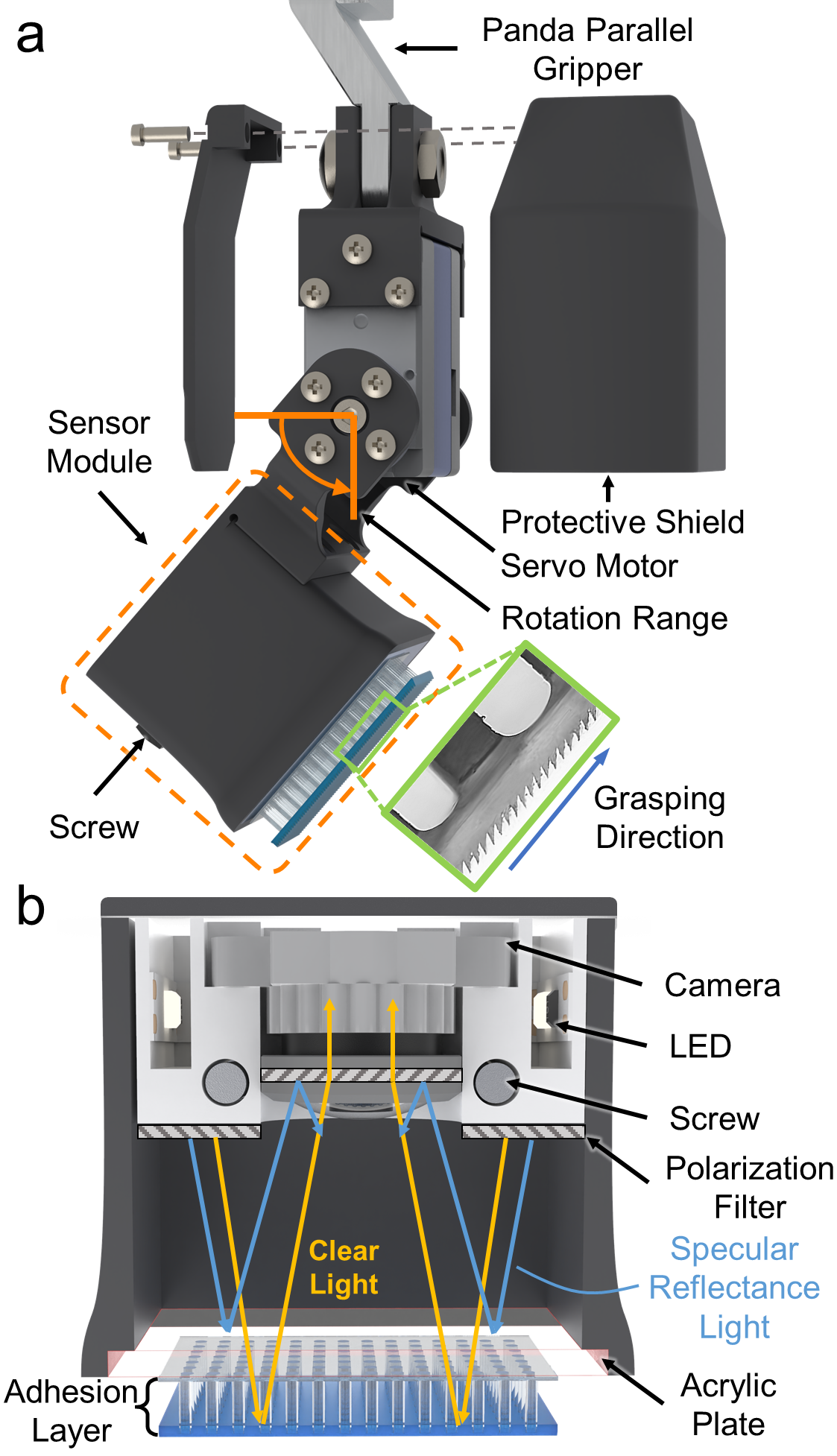}
	\caption{Mechanical design of the adaptive gecko gripper. (a) CAD-rendered side view of the gripper with an exploded view of the protective shield. (b) Cross-section view of the sensor module. The polarization filter's position and orientation are labeled to illustrate how it filters out reflected light.}
	\vspace{-0.4cm}
	\label{fig: design}
\end{figure}

The overall design of the sensing module constitutes a zero-distortion camera, LEDs, housing, acrylic plate, and hierarchical adhesive patch, as illustrated in Fig. \ref{fig: design}(b). The camera is mounted on the inner white resin part and surrounded by LEDs. The adhesive layer adheres to the acrylic plate and is then inserted into the black exterior housing and bonded by Sil-Poxy (Smooth-On).
In the previous study, there existed illumination issues, including uneven lightening at the sensor's corner and reflective light from the acrylic plate. These problems reduce image quality, which has a detrimental effect on the sensor's stability and accuracy.

Herein, the illumination system of the sensor has been redesigned to obtain an uniform luminance and eliminate the reflective light in a compact design. The LEDs are arranged in a circular pattern in the white resin component above the acrylic plate to average the luminance of the sensing areas. The white resin part scatters and evenly spreads the light that passes through it. The polarization filter is used to minimize reflected light from the acrylic plate. The horizontal polarization filter at the bottom of the white resin part converts unpolarized light from the LED to horizontally polarized light. A second polarization filter is placed at a 90$^\circ$ angle to the first filter in the camera's lens to filter out the specular reflectance light from the acrylic palte surface, as illustrated in Fig. \ref{fig: design}(b). With the specular reflectance light filtered out, the camera can capture the adhesion layer's deformation in more detail, resulting in steadier and more accurate contact information.

\section{Sensing Pipeline}
\label{sec: sensing}
Tactile information is essential for an adaptive gecko-inspired gripper to establish a firm grasp. The gripper constitutes three main systems that acquire contact information: learning-based contact area prediction, shear force estimation by a displacement vector field, and incipient slip detection. 
\subsection{Learning-based Contact Area Prediction}
\subsubsection{\textbf{Dataset}} 
A dataset was constructed, consisting of data on two main categories of objects: daily objects and four standard shapes (cross, circle, hexagon, and rectangular). The daily-objects data were collected by pressing objects against the sensor with a random orientation and manual labeling the contact area. The 3D-printed standard shape objects were affixed to a 3-axis motorized stage whose position was precisely controlled by the program. By indenting the object with a random depth ranging from 0.1 to 1 mm and sliding the stage in the X and Y directions with a 2 mm step size, a sequence of 100 images was obtained for each shape. A label was automatically generated according to the object's position and shape.

Because of variance in the fabrication of the sensors, no two sensors were identical. We also added 10 non-contact images of two sensors to the dataset to verify the generalization ability and enhance the robustness of the training model. The final dataset contained approximately 1000 images, divided into training, validation, and testing sets in the ratio 7:2:1, respectively.
\begin{table}[htbp]
\centering
\captionsetup{justification=centering}
\caption{Performance of DeepLabV3+, with five different encoders, in estimating the contact area of the testing set. }
\label{tab:encoder}
    \begin{tabular}{cccc}
        \toprule
        \textbf{Encoder}&\textbf{Params}&\textbf{IoU (\%)}&\textbf{FPS}\\
        \midrule
        MobileNet V2 \cite{sandler2018mobilenetv2}&\textbf{2M}&78.64&\textbf{24}\\
        ResNet-18&11M&78.83&22\\
        ResNet-50&23M&78.34&21\\
        ResNet-101&42M&\textbf{78.89}&16\\
        Se-ResNeXt 50 (32x4d)&25M&78.76&20\\
        \bottomrule
    \end{tabular}
\end{table}
\subsubsection{\textbf{Training and Architecture}} 
DeepLabV3+ \cite{chen2018encoder} was used as the image segmentation model architecture for contact area prediction. It combines a spatial pooling operation and encoder-decoder structure to achieve high performance in recovering segmentation boundary details. To avoid the overfitting problem, data augmentation methods such as random brightness and transformation were used to increase the number of data and enhance the model robustness. The model was trained for 250 epochs using the AdaGrad optimizer with a learning rate of 10$^{-3}$. 
Several encoders with various numbers of parameters were evaluated on the testing set in terms of intersection over union (IoU) and frames-per-second (FPS), as listed in Table \ref{tab:encoder}. The testing results indicate that the encoder's parameter number had a negligible effect on the IoU performance. Thus, MobileNet V2 was chosen as it had the highest FPS.
\subsection{Shear Estimation}
We extracted the tangential displacement of the gecko-inspired adhesive surface by incorporating a red marker pattern on the interior of the hierarchical structure. The markers were extracted from the captured image using thresholding and processed with morphology before blob detection. The vector field was then extracted using a K-D tree by comparing the reference $b_0$ (blob detection mask when there is no contact) with the current mask $b_i$. 
The output was a vector field $U$ of each marker, which was then summed in the X and Y directions for mapping to the shear force $S$ as follows: 
\begin{equation}
    F_s(x)=2.344x-0.1363x^2-0.06845x^3,
    \label{shear_map}
\end{equation}
where $x$ is the sum of the vector field in the X or Y direction.

\begin{figure}[ht]
	\centering
	\includegraphics[width=0.45\textwidth]{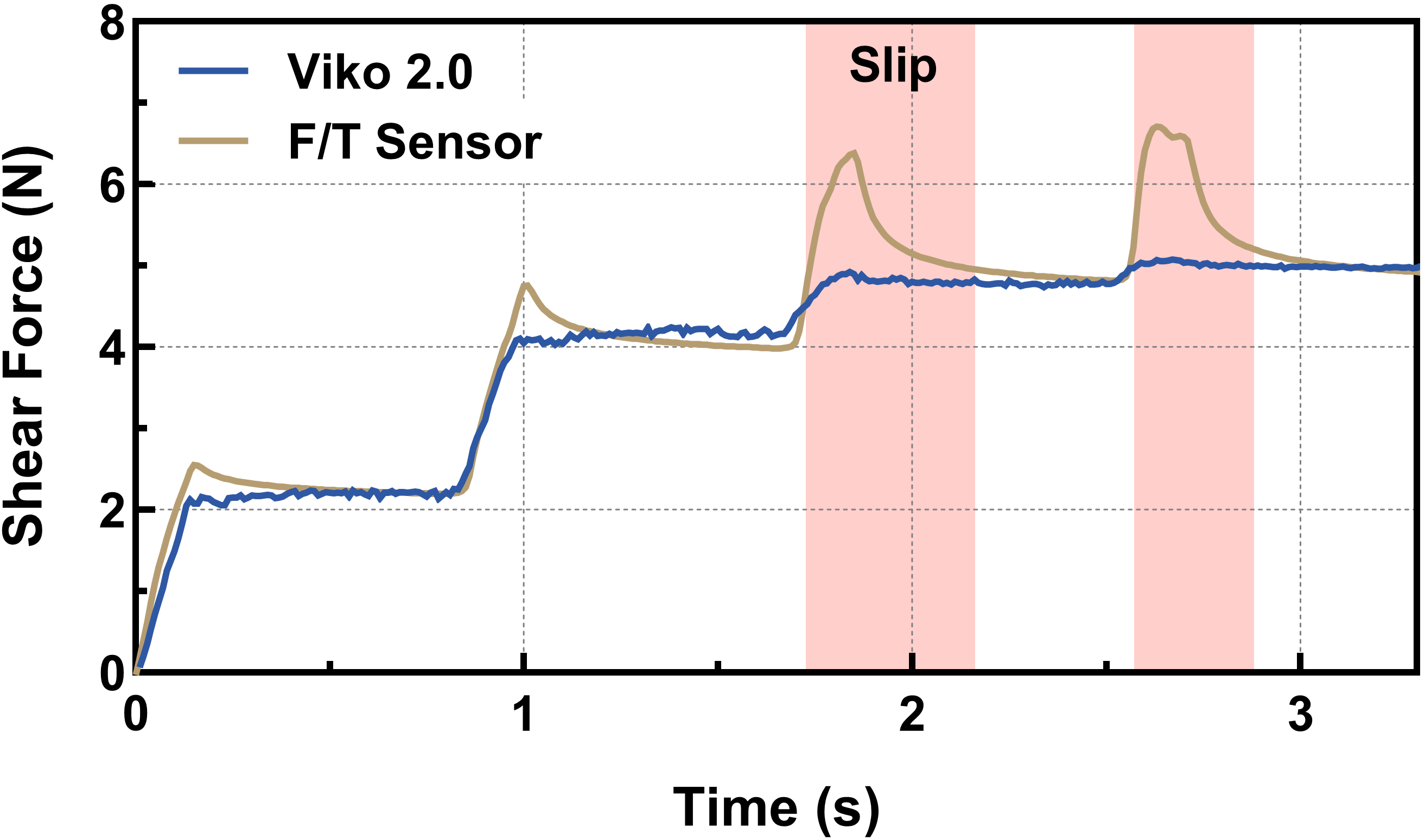}
	\vspace{-0.1cm}
	\caption{The F/T sensor and calibrated \textit{Viko} 2.0 sensing module for shear force measurement showing a close agreement in output.}
	\vspace{-0.4cm}
	\label{fig: shearmapping}
\end{figure}

The entire process of shear force estimation is presented in Algorithm \ref{algo}. The mapping is based on calibration with a 6-axis F/T sensor (Nano-17, ATI) on a custom-built stage \cite{Pang2021}. As plotted in Fig. \ref{fig: shearmapping}, the shear force estimated by our sensor is compared with that of the F/T sensor and demonstrates a good match. 

\begin{figure}
	\centering
	\vspace{0.1cm}
	\includegraphics[width=0.45\textwidth]{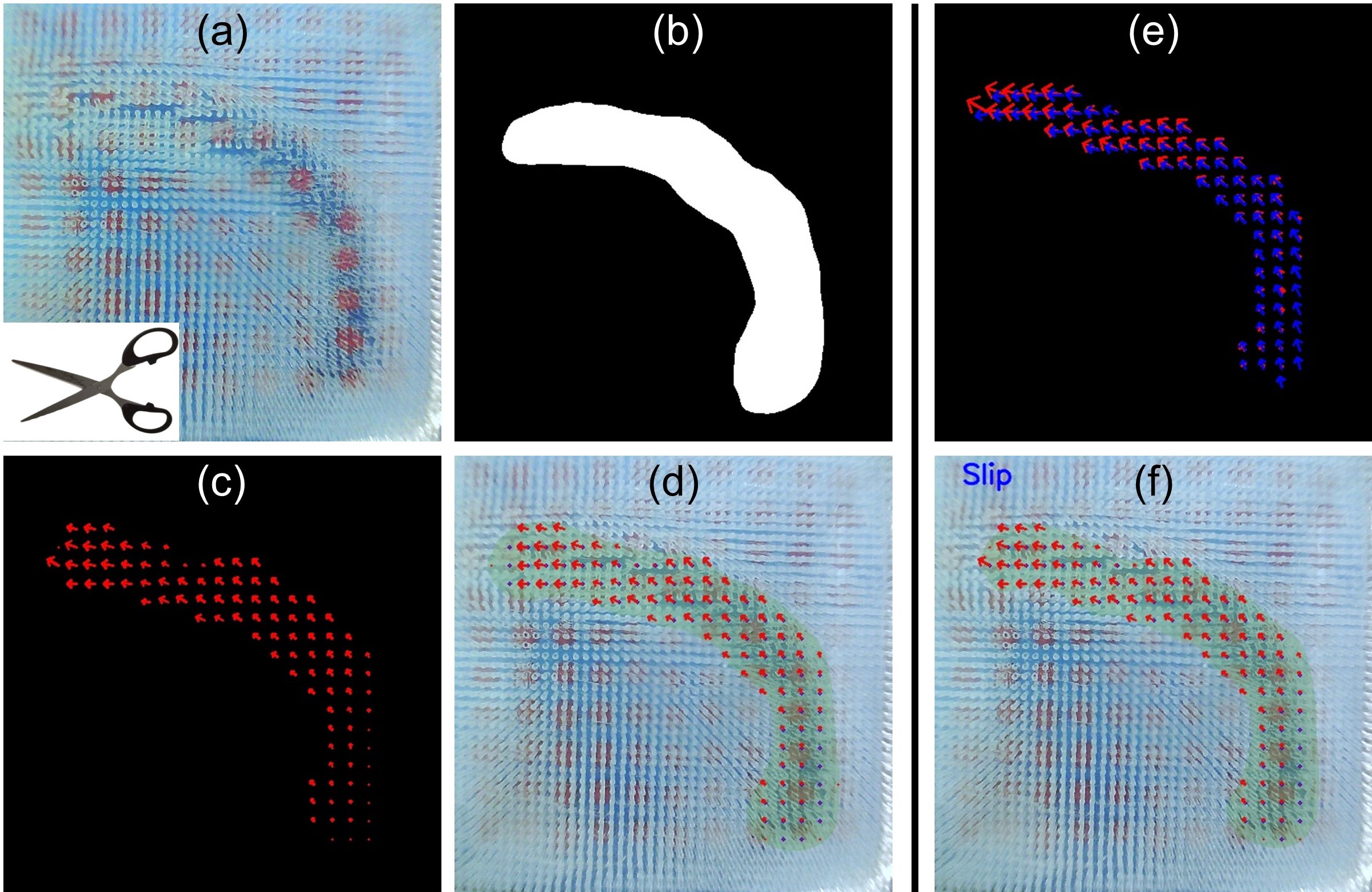}
	\caption{(a) Raw image of a scissor handle (left bottom corner) pressed against the sensor. (b) Mask of the contact area generated by image segmentation. (c) Image of the marker's vector field (red arrow). (d) Processed image with the labels of the contact area (green region) and the marker vector field. (e) The marker's vector field and rigid body motion field (blue arrow). (f) Processed image when incipient slip is detected.}
	\vspace{-0.4cm}
	\label{fig: algo_demo}
\end{figure}

\subsection{Incipient Slip Detection}
The real gecko's adhesive system can slip slowly without a significant drop in shear adhesion \cite{gravish2010rate}, allowing the recovery of a high level of adhesion. Likewise, in a gecko-inspired system, incipient slip detection can prevent grasping failure due to slipping. We followed a similar strategy for incipient slip detection to that proposed by Dong et al. \cite{dong2019maintaining}, who also used a similar sensing marker pattern.

 In stable grasping, the relative position of the object and sensor is constant, and these can be considered a rigid body. Slipping typically occurs when the motion of the contact surface is inconsistent with that of the rigid body. The displacement vector field $U_{i}^{\mathbf{in}}$ of markers inside the contours of the contact area represents the motion of the sensor surface. The rigid body motion field can be computed by assuming that the marker displacement is rigid, and serves as an estimate of the sensor surface motion. When the number of variance between $U_{i}^{\mathbf{in}}$ and $R_i$ exceeds a predefined threshold, an incipient slip is indicated. Figure. \ref{fig: algo_demo}(e) shows the marker's vector field in red arrows and the body motion field in blue arrows, and the corresponding process is shown in Fig. \ref{fig: algo_demo}(f).   
\begin{algorithm}[ht]
\caption{Contact Information Extraction}
\label{algo}
\DontPrintSemicolon
\SetAlgoLined
\KwData{ Initial image $t_0$, \\ ~~~~~~~~~~Image sequence $\Tau =\left \{ t_i | i=1,2,... \right \}$}
\KwResult{Contact area contour $C$, \\~~~~~~~~~~~Shear force $S$, \\ ~~~~~~~~~~~Incipient slip $I$}
$mask_0$ = Morphology(Threshold($t_0$))\;
$b_0$ = Blob detection($mask_0$)\;
\For{$t_i$ \textbf{in} $\Tau$}{
$C_i$ = Image segmentation ($t_i$)\;
$mask_i$ = Morphology(Threshold($t_i$))\;
$b_i$ = Blob detection($mask_i$)\;
\If{Area$(C_i)>0$}{
Displacement vector $U_i$ = KDTree($b_0$,~$b_i$)\;
$S_i$ \,\,~$=$ $F_s$($\textstyle \sum{U_i}$)\;
$b_{0,i}^{\mathbf{in}}$ ~$=$ $C_i$ $\cap$ ~$b_{0,i}$\;
$U_i^{\mathbf{in}}$ $=$ $C_i$ $\cap$ ~$U_i$\;
$R_i$ \,~$=$ rigid body transform($b_0^{\mathbf{in}}$,~$b_i^{\mathbf{in}}$)\;
\If{(number of Threshold($R_i-U_i^{\mathbf{in}}$)) $>$ $6$}{
$I$ $=$ TRUE\;
}
}

}
\end{algorithm}


\begin{figure*}[htbp]
	\centering
	\vspace{0.1cm}
	\includegraphics[width=0.85\textwidth]{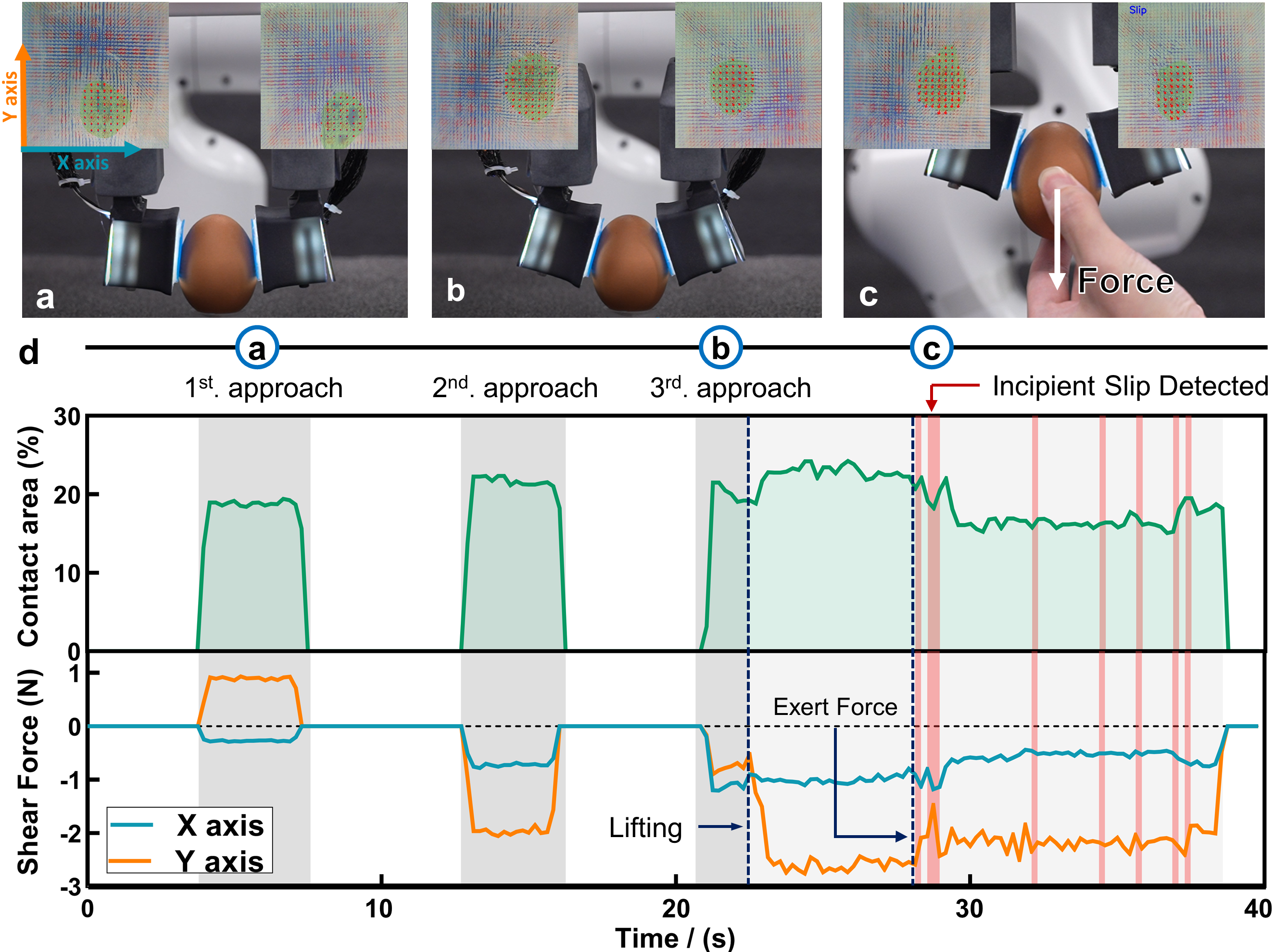}
	\caption{Grasping experiment using the \textit{Viko} 2.0 gripper. Only the sensor on the right jaw of the gripper enables the incipient slip detection for demonstration. Corresponding processed sensor images are depicted at the top (a--c). (a) The gripper first approaches the egg in a parallel configuration. (b) The gripper adjusts the grasping pose, yields enough contact area for grasping the egg, and is ready to pick up the egg. (c) During the grasping process, external downward force is exerted by hand. (d) The sensor's signal chart of total contact area, shear force, and incipient slip (red bar).  }
	\label{fig: exp1}
\end{figure*}
\section{Experiments}
\label{sec: exp}
In practical applications, for gecko-inspired adhesives to have a firm grasp, adequate contact with the target surface is required. Misalignment of the adhesives and the target surface may result in insufficient contact and lead to gripping failure. Although hierarchical structures can compensate for misalignment within a small range, they cannot adapt the surface with large geometrical changes. They can adjust the grasping pose based on the contact information feedback from the sensors to achieve a firmer grasp with a larger contact area. In addition, implementing the incipient slip detection enhances the gripper's robustness when confronted with external disturbances, which is common in pick-and-place tasks. Two experiments demonstrate the gecko-inspired hierarchical adhesive performance and contact information for robust manipulation. The first experiment demonstrates the readjustment of the grasping pose based on the contact area and the maintenance of grasping with incipient slip detection. The second experiment shows the hierarchical structure's high adhesive performance and adaptability in grasping objects with various geometries and surface textures.
As shown in Fig. \ref{fig: intro_graph}, the experiment setup consisted of a robotic arm with a parallel gripper (Franka Emika) and a pair of \textit{Viko} 2.0 fingers. The sensor module captured the image of the hierarchical structure with a 480 $\times$ 480 spatial resolution at 24 Hz. Two cameras in \textit{Viko} 2.0 grippers, and an Arduino board that controlling two servos, were connected to the computer via a USB wire. All computational processes were performed on the computer, with the robot operating system (ROS) network publishing control commands to the servo motors and robotic arm.
\subsection{Sensor-based Grasping}
This experiment evaluated the sensor's functionality in a real-world grasping scenario and demonstrated pose adjustment and incipient slip detection by sensor-based contact information feedback.

The fingertips approached the egg at a parallel mode and obtained a relatively small contact area of 18\%. Then, they re-approached the egg with an adjusted pose at a slightly larger contact angle controlled by the servo motor and achieved a similar contact area of 22\% for the second and third approaches. The gripper can further search for the best grasping pose by re-grasping and varying contact angles. Here, a 22\% contact area was enough to grasp the egg securely. The shear forces during grasping are illustrated in Fig. \ref{fig: exp1}(d), where the sudden increase in shear force indicates that the object has been lifted. An external force was exerted on the egg by hand to simulate the disturbance force during pick-and-place scenarios. When the incipient slip was as shown in fig. \ref{fig: exp1}(c), it sent out the signal for the robotic arm to increase the grasping force to prevent large slippage of the egg. However, some slippage still happened and led to a slight decrease in the contact area. Owing to the sensor's high adhesive force, it still successfully grasped the egg. The decline in shear force indicates placement of the egg, and after the gripper released the egg, the shear force dropped to zero without drifting.

The results show that the contact information feedback of the sensor provides pre-grasp pose evaluation and real-time contact status to enhance the grasping performance. Thus, \textit{Viko} 2.0 combines the benefits of the hierarchical adhesive's high adaptiveness and visuotactile sensor's sensing capabilities.
\begin{figure}[ht]
	\centering
	\includegraphics[width=0.48\textwidth]{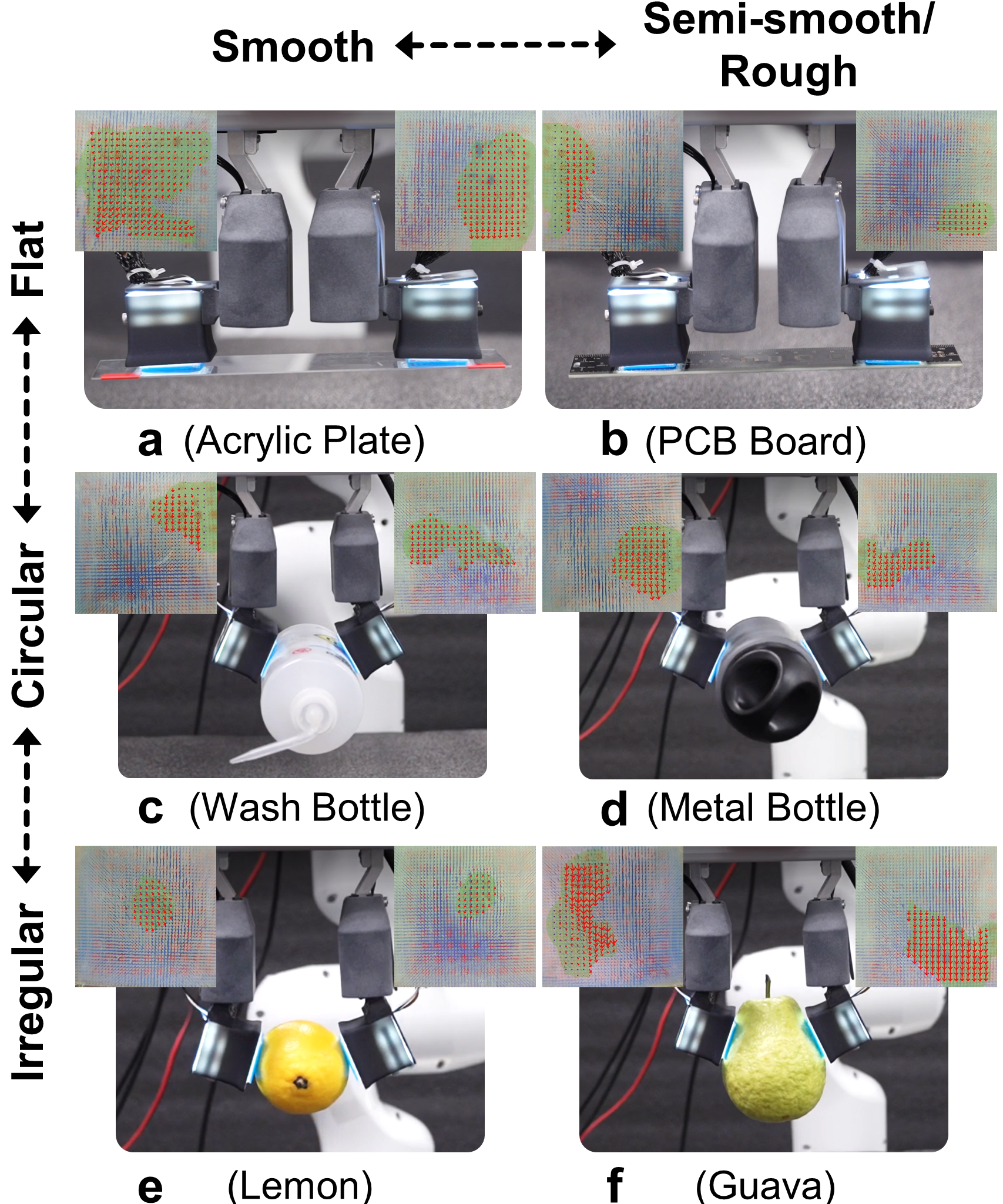}
	\caption{\textit{Viko} 2.0 gripper grasping objects with various geometries and textures. The top of each photo shows the processed image of each sensor during grasping.}
	\label{fig: exp2}
\end{figure}
\subsection{Adaptive Grasping}
We performed a series of experiments to illustrate the gripper's adaptability and robustness to objects with varying geometries and roughness. Figure \ref{fig: exp2} shows photos of objects being grasped and classified into different categories. In the first column, the gripper firmly grasps smooth-surfaced objects. In the second column, the gripper grasps semi-smooth or rough-surfaced objects, which are considered difficult to handle using non-hierarchical adhesives, as the contact area decreases significantly at the micrometer scale.

In Fig. \ref{fig: exp2}(a,b), \textit{Viko} 2.0 grasps thin and flat objects, which can be difficult for parallel grippers without adhesion to grasp, especially if we consider the alignment between two fingers and the object. The hierarchical structure increases the surface alignment tolerance approximately 20 times from 0.1 $mm$ to 2 $mm$, resulting in a successful grasp of the acrylic plate and PCB board with a semi-smooth surface. For circular objects, Fig. \ref{fig: exp2}(c) shows the ability to gently grip a soft wash bottle without deformation, while Fig. \ref{fig: exp2}(d) shows a firm grip on a heavier metal bottle with the same grasping configuration. Figure \ref{fig: exp2}(e,f) shows the handling of irregular-shaped fruits, including a lemon and a guava, with rough and bumpy surfaces. We believe that no previously reported visuotactile gripper is equally capable of all these grasping tasks.

\section{Conclusion and Future Works}
\label{sec: conc}
Gecko-inspired adhesives show potential for robotic grasping and manipulation tasks; however, their applicability has been restricted by the rigorous alignment tolerance and loading conditions required for efficient utilization. The use of hierarchical structures can enhance the alignment tolerance and more evenly distribute the load by passive deformation. However, the complex structure makes sensing difficult in contact states. We present a co-design of hierarchical adhesives and visuotactile sensors in the form of a gripper that can sense and grasp a wide range of object textures and geometries while in the contact state.

Compared with a non-hierarchical structure, a hierarchical structure, comprising materials with different hardness, doubles the contact area and has a 1.5 times higher performance in normal adhesion. For the sensibility of the adhesives, we develop an internal marker array stamping method that allows the sensing module to be attached to the inner surface of the hierarchical structure for vision-based feature tracking. Image segmentation and feature tracking techniques utilize real-time measurements of the contact area, shear force, and incipient slip. We demonstrate adaptation of the grasping pose based on the contact area for more efficient use of the adhesives and detection of the incipient slip under external disturbance. We also test the grasping of general objects, thus showing the ability of the gripper to pick up flat, circular, and irregular-shaped objects with various surface roughness.


In the future, we will explore the tactile servoing with our gripper, focusing on the maintenance of high contact area and even stress distribution, to achieve active control of the gripper for dynamic in-hand manipulation.


\bibliographystyle{IEEEtran}
\bibliography{reference}

\end{document}